\documentclass[10pt,twocolumn,letterpaper]{article}

\usepackage{iccv}
\usepackage{times}
\usepackage{epsfig}
\usepackage{graphicx}
\usepackage{amsmath}
\usepackage{amssymb}

\usepackage{booktabs}
\usepackage{adjustbox}
\usepackage{utfsym}
\usepackage[accsupp]{axessibility}
\usepackage[pagebackref=true,breaklinks=true,letterpaper=true,colorlinks,bookmarks=false]{hyperref}

\iccvfinalcopy % *** Uncomment this line for the final submission

\def\iccvPaperID{4816} % *** Enter the ICCV Paper ID here

\ificcvfinal\fi

\begin{document}

%%%%%%%%% TITLE
\title{GET: Group Event Transformer for Event-Based Vision}

% \title{Event-Based Vision with Group Event Transformer: A Hierarchical and Cross-Domain Approach}

\author{
    Yansong Peng \and
    Yueyi Zhang\thanks{Corresponding author} \and
    Zhiwei Xiong \and
    Xiaoyan Sun \and
    Feng Wu \and
University of Science and Technology of China\\
{\tt\small pengyansong@mail.ustc.edu.cn, 
\{zhyuey, zwxiong, sunxiaoyan, fengwu\}@ustc.edu.cn
}
% \and
}

\maketitle
\ificcvfinal\thispagestyle{empty}\fi

%%%%%%%%% ABSTRACT
\begin{abstract}
%
% This paper provides a new vision Transformer called Group Event Transformer (GET), that serves as a powerful backbone for event-based vision.
% %
% Prior event-based backbones achieve excellent performance by leveraging state-of-the-art image-based designs. However, they barely focus on explicitly modeling the rich temporal and polarity information of events.
% %
% Therefore, we propose GET, which acts as a hierarchical Transformer that accepts asynchronous events as input. 
% %
% GET embeds events into different \textbf{Groups} according to their time and polarity. 
% %
% With Event Token Embedding, Event Dual Self-Attention, and Group Token Aggregation, GET can effectively guide cross-group communication and group-wise feature aggregation in both the spatial and temporal-polarity domains.
% %
% As the stage deepens, the number of groups is gradually reduced, and the features in each group gain a larger temporal-polarity receptive field, which helps to better characterize the object.
% %
% Experimental results performed on four event-based classification datasets (Cifar10-DVS, N-MNIST, N-CARS, and DVS128Gesture) and two event-based object detection datasets (1Mpx and Gen1) demonstrate that our proposed approach outperforms other state-of-the-art methods.
% Event-based vision has gained attention as a promising alternative to frame-based methods. 
% Event-based vision is a promising alternative to frame-based methods.
% However, 
Event cameras are a type of novel neuromorphic sensor that has been gaining increasing attention.
% Event cameras, as novel neuromorphic sensors, have attracted more and more attention.
Existing event-based backbones mainly rely on image-based designs to extract spatial information within the image transformed from events, 
% resulting in an excessive coupling of temporal, spatial, and polarity information during the feature extraction process.
overlooking important event properties like time and polarity. 
To address this issue, we propose a novel \textbf{Group-based} vision Transformer backbone for Event-based vision, called Group Event Transformer (GET), which decouples temporal-polarity information from spatial information throughout the feature extraction process.
% makes full use of the provided temporal and polarity information of events by decouples them with spatial information. 
Specifically, we first propose a new event representation for GET, named Group Token, which groups asynchronous events based on their timestamps and polarities. Then, GET applies the Event Dual Self-Attention block, and Group Token Aggregation module to facilitate effective feature communication and integration in both the spatial and temporal-polarity domains. After that, GET can be integrated with different downstream tasks by connecting it with various heads.
%can be connected with heads for different downstream tasks. 
 % As the network goes deeper, the number of groups decreases and the features in each group have a larger temporal-polarity receptive field, facilitating the capture of object characteristics.
 We evaluate our method on four event-based classification datasets (Cifar10-DVS, N-MNIST, N-CARS, and DVS128Gesture) and two event-based object detection datasets (1Mpx and Gen1), and the results demonstrate that GET outperforms other state-of-the-art methods. The code is available at  \url{https://github.com/Peterande/GET-Group-Event-Transformer}.
\end{abstract}
\vspace{-3mm}

%%%%%%%%% BODY TEXT
\section{Introduction}
Event cameras are a type of bio-inspired vision sensors that capture per-pixel illumination changes asynchronously. Compared with traditional frame cameras, event cameras offer many merits like high temporal resolution (\textgreater 10K fps), high dynamic range (\textgreater 120 dB), and low power consumption (\textless 10 mW) \cite{survey}. 
Many applications, such as object classification \cite{ActionRecognition1, ActionRecognition2} and high-speed object detection \cite{eyegaze, ObjectDetection1, RED}, can take advantage of this kind of camera, especially when power consumption is limited or in the presence of challenging motion and lighting conditions. 
% Diagram of group number changing across GET. The group number is 6 at the initial stage and 1 at the last stage. 
\begin{figure}[t]
\centering
\includegraphics[width=\columnwidth]{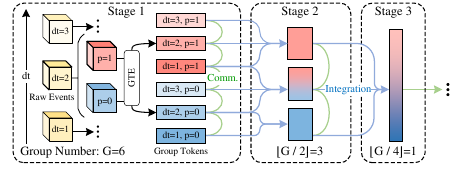}
    \caption{
     Group Tokens are generated in Stage 1 according to timestamps and polarities of events. In GET, these tokens are effectively communicated and integrated, making full use of important event information while maintaining information decoupling. 
     }
    \vspace{-3mm}
\label{fig:Group}
\end{figure}
% $d_{t}$ is discretized time defined in Equation \ref{eq:1}, $p$ is polarity.
% \begin{figure}[t]
% \centering
% \includegraphics[width=\columnwidth]{Figs/Group3.pdf}
% \caption{Group Event Transformer (GET) when initial group number is 6. Different Groups denote different times and polarities. The colored lines indicate the Event Token Embedding, Event Dual Self-Attention, and Group Token Aggregation operations related to the second group with $d_{t}=2$ and $p=1$. ($d_{t}$ is discretized time defined in Equation \ref{eq:1}, $p$ is polarity.)}
% \label{fig:Group}
% \end{figure}
% \begin{figure}[t]
% \centering
% \includegraphics[width=\columnwidth]{Figs/Group2.pdf}
% \caption{Overview of a 3-stage Group Event Transformer (GET). Different Groups denote different times and polarities. The green lines indicate the information flow (feature communication) of the second group with $d_{t}=2$ and $p=1$. ($d_{t}$ is discretized time defined in Equation \ref{eq:1}, $p$ is polarity.)}
% \label{fig:Group}
% \end{figure}

%Grouped Event Tokens}, which are generated by Event Token Embedding (ETE) method. The channels of an Event Token are grouped according to different polarities and times. GET aims to build meaningful cross-group communications.
% 主语Different from traditional cameras, which output three-channel RGB frames at a certain frame rate. 

The event data are stored as asynchronous arrays, containing the location, polarity, and time of each illumination change. In contrast, traditional frames represent visual information as a matrix of pixel values captured at a fixed rate. As a result, traditional image-based neural networks can not be directly applied to event data. 
To address this issue, several works have been proposed to convert events into image-like representations. These representations, such as voxel grid \cite{VOXEL}, event histogram \cite{HIST}, and time surface \cite{HOTS, HATS, Huang_2023_CVPR}, are then fed to deep neural networks \cite{DNN1, DNN3, DNN4, DNN5}. More recent works leverage the successful Transformer architecture to extract features directly from the above event representations \cite{Transformer1, Transformer2, Transformer3, Transformer4}, or from the feature maps obtained by CNN-based backbones \cite{TransformerOpticalFlow, TransformerVideoReconstruction, TransformerDomainAdaptation}. Although these networks have fair performance, they are still limited by their reliance on image-based feature extraction designs \cite{VIT, localvit, TNT, swin}. These designs mainly extract spatial information and can not fully utilize the temporal and polarity information contained in events. There are also some works trying to bridge this gap. For example, Spiking neural networks (SNN) are utilized to capture temporal information by representing event data as spikes that occur over time and propagate through neurons \cite{Wu2021TrainingSN, SNN1, SNN3}. Graph neural networks (GNN) are also adopted to model the dynamic interactions between nodes in an event graph over time \cite{GNN1, GNN2}. However, these two kinds of methods either require specialized hardware or sacrifice performance. 

In this work, we revisit the problem of event-based feature extraction and attempt to fully utilize temporal and polarity information of events with a Transformer-based backbone, while preserving its spatial modeling capability. To this end, we first propose a novel event representation called Group Token, which groups events according to their timestamps and polarities. Then we propose Group Event Transformer (GET) to extract features from Group Tokens. We design two key modules for GET: Event Dual Self-Attention (EDSA) block and Group Token Aggregation (GTA) module. The EDSA block acts as the main self-attention module to extract correlations between different pixels, polarities, and times within Group Tokens. It maintains efficient and effective feature extraction by performing local Spatial Self-Attention and Group Self-Attention and establishing dual residual connections. GTA is placed between two stages to achieve reliable spatial and group-wise token aggregation. It achieves global communication in both the spatial and temporal-
polarity domains by using a novel overlapping group convolution. Figure \ref{fig:Group} shows the overview of a 3-stage GET network with the initial group number of 6. The grouped tokens are fused to produce larger groups, in which the features gradually gain larger spatial and temporal-polarity receptive fields, helping to better characterize objects. 

These novel designs make GET achieve state-of-the-art performance on four event-based classification datasets ($84.8\%$ on Cifar10-DVS, $99.7\%$ on N-MNIST, $96.7\%$ on N-CARS, and $97.9\%$ on DVS128Gesture) and two event-based object detection datasets ($47.9\%$ and $48.4\%$ mAP on Gen1 and 1Mpx). These results all demonstrate our proposed network is beneficial for event-based feature extraction. Our contributions are summarized in the following.

\begin{itemize}
\setlength{\itemsep}{0pt}
\setlength{\parsep}{0pt}
\setlength{\parskip}{0pt}
\vspace{-3mm}
\item We propose a new event representation, called Group Token, that groups asynchronous events based on
their timestamps and polarities.  
\item We devise the Event Dual Self-Attention block, enabling effective feature communication in both the spatial and temporal-polarity domains. 
\item We design the Group Token Aggregation module, which uses the overlapping group convolution to integrate and decouple information in both domains.
\item Based on the above representation and modules, we develop a powerful Transformer backbone for event-based vision, called GET. Experimental results on both classification and object detection tasks demonstrate its superiority.
\end{itemize}

\begin{figure*}[t]
\centering
\includegraphics[width=2\columnwidth]{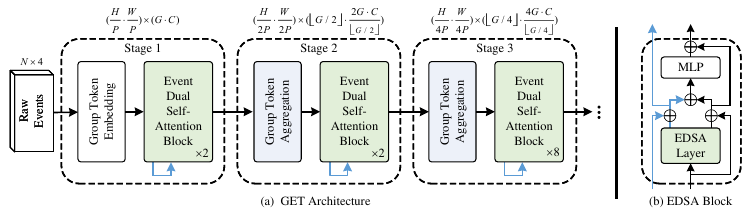}
\caption{
 (a) The architecture of a 3-stage GET; (b) The EDSA block with dual residual connections, blue arrows mark the feature maps after GSA refinement. (All figures omit the normalization and activation layers.) }
 \vspace{-3mm}
\label{fig:Structure}
\end{figure*}
% \vspace{-3mm}
\section{Related Work}
\noindent\textbf{Event Representation.}
As the event streams are asynchronous and sparse, it is necessary to convert them into appropriate alternative representations.
% except for a few filter-based and SNN-based works \cite{filter-based, SNN-based, SNN1, SNN2, SNN3}. 
Image-like event representations are widely utilized, such as voxel grid \cite{VOXEL}, event histogram \cite{HIST} and time surface \cite{HOTS, HATS}. However, these representations discard partial information of events, leading to a degradation in performance. 
There are also works introducing learning-based representations \cite{RecurrentSurface, sparseconvolutional}, but these approaches introduce redundancy and latency to the network.
Additionally, some recent works have accomplished graph-based representations \cite{GNN1, GNN2}, but have yet to yield satisfactory performance.

% \subsection{Event-Based Vision}
% Event cameras offer many advantages over other vision modalities. Benefiting from their high temporal resolution, event cameras are capable of high-speed object detection/tracking \cite{eyegaze, linetracking, ObjectDetection1, RED}. Event cameras are also suitable for surveillance and monitoring \cite{monitor, monitoranimal}, and object/gesture recognition \cite{DNN4, Transformer3}, thanks to their high dynamic range and low power consumption, especially when power consumption is limited or when dealing with challenging motion and lighting conditions.

\noindent\textbf{Self-Attention Mechanism.}
% In deep neural networks, different channels of
% feature maps usually represent different colors, textures, or other characters \cite{channelimportance}. Many excellent channel attention mechanisms try to improve the channel modeling capability of spatial attention. SENet \cite{senet} is the first proposed channel attention mechanism, which has many follow-ups \cite{improvese1, improvese2, improvese3}. However, these mechanisms only generate channel weights that reflect importance rather than integrating channel-wise inter-correlations. As a result, they are unable to explicitly model sequential information such as time. 
Self-attention mechanism \cite{Attentionis, bert} is an integral part of deep neural networks, which was first widely used in the field of natural language processing (NLP). In the computer vision community, Vision Transformer (ViT) \cite{VIT} is a seminal work that contributes an artful vision backbone by applying self-attention to images. Dual attention \cite{dual} introduces a novel self-attention method that extracts channel information and saves it into the spatial domain and greatly improves performance. But it is still limited by the high computational costs of global attention calculation, and its direct fusion operation of spatial and channel feature maps also results in interference. More recently, many variants of ViT \cite{dataefficientvit, t2tvit, TNT, localvit, swin} are proposed to improve performance and data/computation efficiency by introducing the local self-attention operation or building hierarchical Transformer architectures. 

\noindent\textbf{Event-Based Transformer.}
% Transformers, also known as self-attention mechanisms \cite{Attentionis, bert, imagetransformer, DETR}, have achieved remarkable success in various machine learning tasks. In the computer vision community, Vision Transformer (ViT) \cite{VIT} is a seminal work that contributes an artful vision backbone by applying the Transformer encoder to images. Many variants of ViT \cite{dataefficientvit, t2tvit, TNT, localvit, swin} are proposed to improve performance and data/computation efficiency by introducing local operation or building hierarchical structures.
Transformer networks have gained popularity in various event-based tasks, such as classification \cite{EvT}, object detection \cite{Transformer4}, and video reconstruction \cite{TransformerVideoReconstruction}. Some of them employ self-attention operations on feature maps obtained through CNN-based backbones \cite{TransformerOpticalFlow, TransformerDomainAdaptation, TransformerVideoReconstruction}, as global attention mechanisms they adopt are computationally heavy. Recently, some studies have focused on directly extracting features from event representations \cite{Transformer1, Transformer2, Transformer3, Transformer4}. However, their Transformer designs are often unable to make full use of the provided event properties like time and polarity.

%-------------------------------------------------------------------------
\section{Method}
\subsection{Main Architecture}
Our proposed GET is a 3-stage Transformer-based network, the architecture of which is illustrated in Figure \ref{fig:Structure}(a). Three modules are utilized within stages: GTE module, EDSA block, and GTA module. At Stage 1, raw events are embedded into Group Tokens via GTE. Then the tokens are fed to 2 sequential EDSA blocks. At Stage 2\&3, one GTA module is first deployed, followed by 2 and 8 EDSA blocks, respectively. The structure of an EDSA block, which combines an EDSA layer, followed by an MLP layer and  normalization layers, is shown in Figure \ref{fig:Structure}(b). It partitions Group Tokens and applies local self-attention in both spatial and temporal-polarity dimensions. The GTA module, which involves overlapping group convolution followed by max pooling, is employed to integrate and decouple spatial and temporal-polarity information between two stages. Residual connections of feature maps after Group Self-Attention refinement are marked by blue arrows, and there are no cross-stage residual connections. The output feature maps are sent to heads of different downstream tasks like classification and object detection.

\subsection{Group Token Embedding}
The Group Token Embedding (GTE) module is designed to convert an event stream to group tokens. The event streams from an event camera with the $H\times W$ resolution, are described as  $(\vec{t}, \vec{p}, \vec{x}, \vec{y})$, where $\vec{t}$, $\vec{p}$ and $(\vec{x}, \vec{y})$ are the time, polarity and location coordinates. 
%Group Tokens are a 2D representation with shape $(\frac{H}{P} \cdot \frac{W}{P}) \times (G\cdot  2P^{2})$, where $G$ is the number of groups splitting $K$ time intervals with two different polarities, $P$ is the patch size, $H$ and $W$ are height and width. A single Group Token consists of $G$ groups of patches. 
We first discretize the time of asynchronous events to $K$ intervals and encode location coordinates into the rank of $P\times P$ patches and positions of patches. We also define discretized time $\vec{d_{t}}$, patch rank $\vec{pr}$, and patch position $\vec{pos}$ as:
\begin{equation}
\left\{\begin{matrix}
\vec{d_{t}} &= &\lfloor K \times (\vec{t}-t_{0})\ /\  (t_{end}-t_{0}+1) \rfloor \hfill\\ 
\vec{pr} &= &\lfloor \vec{x}\mod{P} \rfloor + \lfloor \vec{y}\mod{P} \rfloor \times P \hfill\\ 
\vec{pos} &= &\lfloor \vec{x} \ /\  P \rfloor + \lfloor \vec{y} \ /\  P \rfloor \times (W / P) \hfill.
\end{matrix}\right.
\label{eq:1}
\end{equation}
Then, we map the polarity array and the above three arrays to a single 1D array, which can be described as:
\begin{equation}
\vec{l}=(K\cdot H\cdot W) \cdot\vec{p}+(H\cdot W)\cdot\vec{d_{t}}+(\frac{H\cdot W}{P^{2}}) \cdot \vec{pr} + \vec{pos}.
\end{equation}
Using 1D bin count operation with weights of $\vec{1}$ and relative time $(\vec{t}-t_{0})/(t_{end}-t_{0})$, two 1D arrays with length $H \cdot W \cdot 2K$ are generated. After concatenation and reshape operations, we get event representations with the shape $(\frac{H}{P} \cdot \frac{W}{P}) \times (2K\cdot  2P^{2})$. 

Early convolutions have been proven effective for Transformer networks \cite{earlyconv, CoAtNet}. Thus we further embed the representations with a $3 \times 3$ group convolution layer following an MLP layer. Finally, we get the Group Tokens with the dimension $ \left( \frac{H}{P}\cdot\frac{W}{P} \right) \times \left(G \cdot  C \right)$, where $C$ indicates the channel number of each group. The variable $G$ is the number of groups with different combinations of time interval and polarity. In other words, $G$ is equal to either $2K$ or $2\cdot \frac{K}{2}$, depending on the group division of the convolution layer.
% , where the embed dimension $G \cdot  C$ is a hyper-parameter controlling the model size. 
% More details are provided in the Appendix. 

\begin{figure}[t]
\centering
\includegraphics[width=\columnwidth]{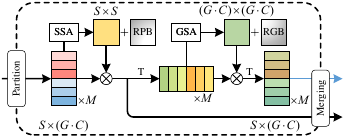}
\caption{
    \textbf{Event Dual Self-Attention process.} SSA is first performed on the partitioned tokens. GSA is then applied. RPB and RGB are added to two attention maps.}
 \vspace{-3mm}
\label{fig:DA}
\end{figure}

\subsection{Event Dual Self-Attention Block}
% As illustrated in Figure \ref{fig:Structure}(b), the black arrows mark the main branch feature maps and the blue arrows mark the feature maps after channel self-attention refinement. They have independent residual connections inside an EDSA block. These connections avoid spatial and channel attention interference, resulting in stable performance improvements. After residual summations, the two branches are summed together and pass an MLP layer to share information. The output of the MLP layer acts as the new main branch feature maps in flowing layers. 
The EDSA block is composed of an EDSA layer, an MLP layer, and normalization layers, as shown in Figure \ref{fig:Structure}(b). It is designed to effectively extract features in both the spatial and temporal-polarity domains. However, as the contextual information in two domains varies, the direct fusion of two kinds of feature maps usually causes interference and biases the network to a local optimum. To address this issue, the EDSA block employs dual residual connections, as shown by black and blue arrows. These dual residual connections avoid information loss in a certain domain by retaining information from previous layers separately in both domains.
% Dual residual connections avoid the features extracted by GSA interfering with those extracted by SSA. 
% The EDSA achieves effective cross-pixel and cross-group communications. 

Figure \ref{fig:DA} shows the structure of an EDSA layer that is included in an EDSA block. For efficient computation,  the first step in the EDSA layer is partitioning the feature maps into $M$ non-overlapping windows with size $S$ to compute parallel self-attention, as in Swin-Transformer \cite{swin}. Then, Spatial Self-Attention (SSA) is performed on each window with shape $S \times (G\cdot C)$ to capture spatial contextual information. The SSA operation can be described as:
\begin{equation}
\begin{split}
&Q,K,V= XW^Q, XW^K, XW^V\\
%\in \mathbb{R}^{S\times (G\cdot C)}
&\text{SSA}(Q,K,V)=\text{Softmax}(\frac{QK^{T}}{\sqrt{G\cdot C}}+B_p)V ,
\end{split}
\end{equation}
where $W^Q, W^K, W^V\in\mathbb{R}^{(G\cdot C) \times (G\cdot C)}$ are linear weights to transform input into query $Q$, key $K$, and value $V$ along the channel dimension. $B_p$ is the relative position bias (RPB), as in \cite{rpb, rpb2}. The attention map is generated by a dot product between $Q$ and $K$. The feature maps are yielded by a matrix multiplication operation between the attention map and the value $V$.

After performing SSA, we transpose the output feature maps into shape $(G\cdot C)\times S $ to perform group self-attention (GSA). Our proposed GSA captures temporal and polarity contextual information of the Group Tokens to refine the semantic features. The GSA operation can be described as:
\begin{equation}
\begin{split}
&Q_g,K_g,V_g= X^TW^Q_g, X^TW^K_g, X^TW^V_g \\ 
%\in \mathbb{R}^{(G\cdot C)\times S}
&\text{GSA}(Q_g,K_g,V_g)=\text{Softmax}(\frac{Q_gK_g^{T}}{\sqrt{S}}+B_g)V_g ,
\end{split}
\end{equation}
where $W^Q_g, W^K_g, W^V_g\in \mathbb{R}^{S \times S}$ are linear weights to transform input along the spatial dimension. We include a relative group bias (RGB) $B_g \in \mathbb{R}^{(G\cdot C)\times S}$ to guide the attention process. The indexes of relative group bias are between $-G+1$ and $G-1$, denoting different feature groups. The attention map is generated by a dot product between $Q_g$ and $K_g$. A matrix multiplication operation between the attention map and $V_g$ calculates the feature maps. 

Finally, the EDSA layer outputs two feature maps: one obtained by applying SSA only (black arrow), and the other obtained by applying both SSA and GSA (blue arrow). The EDSA block guarantees effective feature communication in both
the spatial and temporal-polarity domains.

\subsection{Group Token Aggregation}
GTA is a novel approach that we use to address the limitations of existing token aggregation methods in hierarchical Transformers. To maintain translational equivariance, well-organized token aggregation is important, as highlighted by previous works \cite{nonlocal, nest}. For example, Swin-Transformer \cite{swin} and Nested-Transformer \cite{nest} use shift-window and conv-pool methods to integrate spatial information into the channel domain. However, these methods destroy the group correlations and are unable to integrate information in the temporal-polarity domain. In contrast, GTA utilizes a new convolution method called overlapping group convolution, which effectively integrates and decouples information in both domains.

\begin{figure}[t]
\centering
\includegraphics[width=\columnwidth]{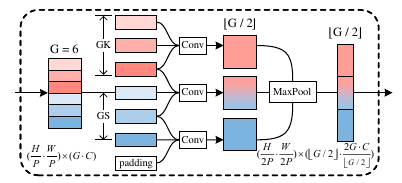}
\caption{
\textbf{Group Token Aggregation process.} We achieve reliable spatial and temporal-polarity information integration by proposing overlapping group convolution.}
\label{fig:GTA}
 \vspace{-3mm}
\end{figure}

Overlapping group convolution is designed based on the concept of group convolution \cite{groupconv}. As illustrated in Figure \ref{fig:GTA}, the input feature maps of overlapping group convolution have shape $(\frac{H}{P} \cdot \frac{W}{P}) \times (G\cdot C)$. It is then divided into $\lfloor G/2 \rfloor$ new overlapping groups and convolved with a set of kernels that have been partitioned into the same number of groups $\lfloor G/2 \rfloor$. GTA has two parameters: The group-wise kernel (GK) denotes the input group number of each kernel and the group-wise stride (GS) denotes the group-wise distance between two inputs. In GET, GK and GS are chosen as $min(\lfloor G/2\rfloor+1, 3)$ and $min(\lceil G/2\rceil - 1, 2)$. A padding with shape $(\frac{H}{P} \cdot \frac{W}{P}) \times C$ is added if needed.

The GTA module consists of a $3\times3$ overlapping group convolution, followed by a normalization layer and a $3\times3$ max pooling operation, where $3\times3$ denotes the kernel size. After using overlapping group convolution, the channel number is doubled, and the group number is halved. The features in each group gain a larger ($3 \times$) receptive field in both the spatial and temporal-polarity domains. This gradual inter-group integration is similar to inter-pixel integration in CNN networks, where feature maps gain a larger receptive field as the network deepens. Moreover, the multi-kernel design of overlapping group convolution can reduce computational cost and model size without compromising performance \cite{groupconv}. The max pooling further down-samples the tokens. This results in a final output shape of $(\frac{H}{2P} \cdot \frac{W}{2P}) \times (\lfloor G/2 \rfloor \cdot \frac{2G\cdot C}{\lfloor G/2 \rfloor})$. Through ablative experiments, it is proved that GTA achieves reliable information integration in both the spatial and temporal-polarity domains.

\begin{table*}[t!] %[htbp]\footnotesize
	\centering
        \begin{tabular*}{\hsize}{@{}@{\extracolsep{\fill}}lcccccc|}

        & \multicolumn{1}{c}{} 
        & \multicolumn{1}{c}{Cifar10-DVS} 
        & \multicolumn{1}{c}{N-MNIST}
        & \multicolumn{1}{c}{N-CARS} \\ 
        \toprule \toprule
        Methods\
        & \multicolumn{1}{c}{Input Type}
        & \multicolumn{1}{c}{top-1 acc. (\%)} 
        & \multicolumn{1}{c}{top-1 acc. (\%)} 
        & \multicolumn{1}{c}{top-1 acc. (\%)}
        & \multicolumn{1}{c}{Params (M)}  \\ \midrule

        HATS \cite{HATS}
        & \multicolumn{1}{c}{Frame} 
        & \multicolumn{1}{c}{52.4}
        & \multicolumn{1}{c}{99.1}
        & \multicolumn{1}{c}{90.2}
        & \multicolumn{1}{c}{-}  \\   
        
        AMAE \cite{AMAE}
        & \multicolumn{1}{c}{Frame} 
        & \multicolumn{1}{c}{62.0}
        & \multicolumn{1}{c}{98.3}
        & \multicolumn{1}{c}{93.6}
        & \multicolumn{1}{c}{21.8}  \\

        MVF-Net \cite{MVF-Net}
        & \multicolumn{1}{c}{Frame} 
        & \multicolumn{1}{c}{55.8} 
        & \multicolumn{1}{c}{98.1}
        & \multicolumn{1}{c}{92.7}
        & \multicolumn{1}{c}{33.5}  \\ 
        
        RG-CNNs \cite{RGCNNS}
        & \multicolumn{1}{c}{Voxel} 
        & \multicolumn{1}{c}{54.0} 
        & \multicolumn{1}{c}{99.0}
        & \multicolumn{1}{c}{91.4}
        & \multicolumn{1}{c}{19.5}  \\ 
        
        EV-VGCNN \cite{VGCNN}
        & \multicolumn{1}{c}{Voxel} 
        & \multicolumn{1}{c}{65.1} 
        & \multicolumn{1}{c}{99.4}
        & \multicolumn{1}{c}{95.3}
        & \multicolumn{1}{c}{0.8}  \\ 
        
        VMV-GCN \cite{VMVGCN}
        & \multicolumn{1}{c}{Voxel} 
        & \multicolumn{1}{c}{66.3} 
        & \multicolumn{1}{c}{99.5}
        & \multicolumn{1}{c}{93.2}
        & \multicolumn{1}{c}{0.9}  \\   \midrule
        
        DT-SNN \cite{DTSNN}
        & \multicolumn{1}{c}{Spike} 
        & \multicolumn{1}{c}{60.5} 
        & \multicolumn{1}{c}{99.5}
        & \multicolumn{1}{c}{-}
        & \multicolumn{1}{c}{-}  \\     
        
        PLIF \cite{PLIF}
        & \multicolumn{1}{c}{Spike} 
        & \multicolumn{1}{c}{74.8} 
        & \multicolumn{1}{c}{99.6}
        & \multicolumn{1}{c}{-}
        & \multicolumn{1}{c}{4.7 - 17.1}  \\
        EvS \cite{EvS}
        & \multicolumn{1}{c}{Graph} 
        & \multicolumn{1}{c}{68.0} 
        & \multicolumn{1}{c}{99.1}
        & \multicolumn{1}{c}{93.1}
        & \multicolumn{1}{c}{-}  \\ \midrule
        
        % Event Transformer \cite{EventTransformer}
        % & \multicolumn{1}{c}{token} 
        % & \multicolumn{1}{c}{71.2} 
        % & \multicolumn{1}{c}{\textbf{99.9}}
        % & \multicolumn{1}{c}{95.4}
        % & \multicolumn{1}{c}{15.9}  \\ 

        Swin-T v2 \cite{swinv2}
        & \multicolumn{1}{c}{Token} 
        & \multicolumn{1}{c}{77.2}
        & \multicolumn{1}{c}{99.3} 
        & \multicolumn{1}{c}{92.8}
        & \multicolumn{1}{c}{6.9}  \\
        Nested-T \cite{nest}
        & \multicolumn{1}{c}{Token} 
        & \multicolumn{1}{c}{78.1}
        & \multicolumn{1}{c}{99.3} 
        & \multicolumn{1}{c}{93.3}
        & \multicolumn{1}{c}{4.2}  \\
        Ours
        & \multicolumn{1}{c}{Token} 
        & \multicolumn{1}{c}{\textbf{84.8}}
        & \multicolumn{1}{c}{\textbf{99.7}} 
        & \multicolumn{1}{c}{\textbf{96.7}}
        & \multicolumn{1}{c}{4.5}  \\

        \bottomrule
	\end{tabular*}%
        \vspace{1mm}
	\caption{Classification performance on the Cifar10-DVS, N-MNIST, and N-CARS datasets.}
 \vspace{-3mm}
	\label{tab:classification}%
\end{table*}%

\section{Experiments}
We first demonstrate the comparison results of GET and other state-of-the-art works on four event-based classification datasets and two event-based object detection datasets. Then we present ablative experimental results on both tasks to analyze the major contributions of the proposed representation and modules. Finally, we show the visualizations of feature maps and detection results. 
\subsection{Experimental Setup}
\noindent\textbf{Datasets.}  
We evaluate our proposed methods on four event-based classification datasets, which are Cifar10-DVS \cite{CIFAR10-DVS}, N-MNIST \cite{nmnist}, N-CARS \cite{HATS}, and DVS128Gesture \cite{dvs128gesture}. We also test the methods on two representative large-scale object detection datasets, which are the Gen1 dataset \cite{gen1} and the 1Mpx dataset \cite{RED}.

The Cifar10-DVS dataset contains 10,000 event streams converted from the up-sampled images in the Cifar10 dataset \cite{cifar}. The resolution of these streams is $128\times128$ and there are 10 different classes. The N-MNIST dataset includes 70,000 event streams converted from the original MNIST dataset \cite{mnist}, which have 10 different classes and a resolution of $34\times34$. The N-CARS dataset includes 12,336 car samples and 11,693 non-car samples. The resolutions of samples in N-CARS are different. The highest resolution is not higher than $128\times128$. The DVS128Gesture dataset contains 1,342 real-world event streams collected by a $128\times128$ event camera. There are 11 different gesture classes (including a class of random movements). 

The Gen1 dataset is the most commonly used event-based object detection dataset, which consists of 39 hours of events. 
%Within the dataset, 22.63 hours are allocated for training, 6.59 hours for validation, and 10.10 hours for testing. 
The Gen1 dataset is collected with a resolution of $304\times240$ and contains 2 object classes. The labeling frequency of the Gen1 dataset is 20 Hz. Another dataset, the 1Mpx dataset contains 14.65 hours of events.
%, with 11.19 hours for training, 2.21 hours for validation, and 2.25 hours for testing. 
The resolution of samples in the 1Mpx dataset is $1280\times720$. The labeling frequency of this dataset is 60 Hz, and the number of bounding boxes exceeds 25M. There are 7 labeled object classes. Following \cite{ASTM, RED}, we only utilize 3 classes (car, pedestrian, and two-wheeler) for performance comparison.

\noindent\textbf{Implementation Details.}
For classification experiments, we follow the official train/test split for N-MNIST, N-CARS, and DVS128Gesture datasets for fair comparisons. For the Cifar10-DVS dataset, we follow the split adopted by previous studies \cite{DTSNN, PLIF} to select samples for testing. 
% , wherein $1/10$ samples are selected for testing.
Our approach applies random horizontal flip and random spatial-temporal crop augmentations on raw events to classification tasks. Since the DVS128Gesture dataset includes both right and left-hand movements, target transformation is applied after using a horizontal flip. Random horizontal flip augmentation is not used on the N-MNIST dataset. We also include Mixup \cite{mixup} and Cutmix \cite{cutmix} to overcome the over-fitting problem on certain datasets. We train our classification models from scratch for 1000 epochs. 

For object detection experiments, the train/test split of Gen1 and 1Mpx datasets are predefined. We choose the most commonly used 50 ms as the time interval of each sample and follow the dataset processing of previous works \cite{Transformer4, RED, ASTM} to remove small bounding boxes and downsample events. 
% The bounding boxes of the 1Mpx dataset with a side length smaller than 20 pixels and a diagonal size smaller than 60 pixels are removed. And the bounding boxes of the Gen1 dataset with a side length smaller than 10 pixels and a diagonal size smaller than 30 pixels are removed. 
% We also downsample the sample resolution of the 1Mpx dataset to $640\times360$. 
We follow the same training strategy as RVT \cite{Transformer4}.
Our detection models are trained from scratch for 400000 steps. 

\begin{table}[t]
  \centering
    \tabcolsep=0.088cm
    % \begin{adjustbox}{width=\columnwidth,center}
    \begin{tabular*}{\hsize}{@{}@{\extracolsep{\fill}}lccc@{}}
    
    % & \multicolumn{1}{c}{} 
    & \multicolumn{3}{c}{DVS128Gesture} \\ \toprule \toprule
    Methods
    & \multicolumn{1}{c}{Input Type} 
    & \multicolumn{1}{c}{top-1 acc. (\%)}
    & \multicolumn{1}{c}{Params (M)}\\  \midrule
    PLIF \cite{PLIF}
    &  Spike
    &  97.6
    &  1.7 \\ 
    SLAYER \cite{SLAYER}
    &  Spike
    &  93.6
    &  - \\ 
    TORE \cite{tore}
    &  Tore
    &  96.2
    &  5.9 \\ 
    PointNet++ \cite{pointnet}
    &  Clouds
    &  95.3
    &  3.5 \\ \midrule
    EvT \cite{EvT}
    &  Token
    &  96.2
    &  0.5 \\
    Swin-T v2
    \cite{swinv2}
    &  Token
    &  93.2
    &  7.1 \\
    Nested-T \cite{nest}
    &  Token
    &  93.8
    &  4.2 \\
    Ours
    &  Token
    &  \textbf{97.9}
    &  4.5 \\

    \bottomrule
    \end{tabular*}%
    % \end{adjustbox}
  \vspace{1mm}
  \caption{Classification performance on the DVS128Gesture dataset, which includes long time-range samples.}
   \vspace{-3mm}
  \label{tab:gesture}%
\end{table}% 

We use a 3-stage GET with an embedding dimension of 48 on classification tasks. The group number $G$ of Cifar10-DVS, N-MNIST, N-CARS, and DVS128Gesture datasets are chosen as 12, 12, 12, and 24, respectively. For object detection tasks, we integrate a 4-stage GET with an embedding dimension of 72 into RVT \cite{Transformer4}. The $G$ chosen for Gen1 and 1Mpx datasets is 12. We utilize 8 NVIDIA Tesla A800 GPUs for training. When comparing the running time, we only use one NVIDIA GTX 1080Ti GPU.

\subsection{Experimental Results}
\noindent\textbf{Classification.}
We compare GET with other state-of-the-art methods on Cifar10-DVS, N-MNIST, N-CARS, and DVS128Gesture datasets and report the results in Table \ref{tab:classification} and Table \ref{tab:gesture}. Swin-Transformer v2 (Swin-T v2) \cite{swinv2} and Nested-Transformer (Nested-T) utilizing patched voxel grid \cite{VOXEL} as input are also included in the comparison. The embedding dimension of them is also chosen as 48. The time bin number of the voxel grid they used ranges from 4 to 48, but we only list their optimal results. We utilize top-1 accuracy (top-1 acc.) to evaluate the classification performance. We also provide parameter numbers (Params) of the models for a fair comparison. 

On the Cifar10-DVS dataset, GET outperforms other methods by a large margin ($84.8 \%$ vs. $74.8 \%$, PLIF). Compared with the Nested-T, its top-1 accuracy also increases by $6.7\%$. On the N-MNIST dataset, GET reaches the highest performance ($99.7 \%$ vs. $99.6 \%$, PLIF). Compared with Nested-T, its top-1 accuracy is increased by $0.4\%$. On the N-CARS dataset, there is a significant performance gain using our method ($96.7 \%$ vs. $95.3 \%$, EV-VGCNN). Its top-1 accuracy is increased by $3.4\%$ compared with Nested-T.

% Table \ref{tab:gesture} provides the results on the DVS128Gesture dataset. The samples in the DVS128Gesture dataset are long-time-range event streams. Unlike samples in the other three datasets, these event streams are more similar to videos than images. SNN-based networks are more suitable for this kind of dataset. When using large time stamps, 
On the DVS128Gesture dataset, which contains long-time-range event streams.
SNN-based methods \cite{STSC-SNN, PLIF, ActionRecognition2, SLAYER} split an event stream into T bins and feed them iteratively into the network. The inference process is executed T times. As a result, many works \cite{tore, EvT} refrain to compare with the SNN-based works on this dataset. 
% SNN-based networks are more suitable, especially when the time window of SNN is large. Other works \cite{tore, EvT} often refrain to compare with the SNN-based works on this dataset. 
However, the encouraging results in Table \ref{tab:gesture} indicate that GET achieves better performance ($97.9 \%$ vs. $97.6 \%$, PLIF) with only a single-pass inference. This means that our methods can better learn the long-range temporal information of event samples without introducing heavy 3D / recurrent designs.

\noindent\textbf{Object Detection.}
We further evaluate GET on two large-scale event-based object detection datasets, which are the Gen1 dataset and the 1Mpx dataset. The results compared with state-of-the-art works are reported in Table \ref{tab:Gen1} and Table \ref{tab:1MPX}. We use the COCO mean average precision (mAP) \cite{COCO} as the main metric. We also provide the comparison on running time (runtime), which includes the event conversion time
and the network inference time. 

The labeled bounding boxes of the Gen1 dataset and the 1Mpx dataset are generated by detection on frames, thus many areas with few events may still be labeled, since event cameras only capture moving objects and objects with illumination changes. Therefore, some works use LSTM or other memory mechanisms to enhance performance \cite{Transformer4, RED,ASTM, liu2022eventbased}. To comprehensively evaluate the capability of the networks, we compare the performance in two scenarios. The first scenario is that the comparison networks don't utilize memory modules for enhancement. The second scenario is the opposite. 
% Such results can provide a more accurate assessment of the underlying feature extraction capabilities. 
For GET, we use the YOLOX framework \cite{yolox2021} and place the ConvLSTM layers \cite{ConvLSTM} at the backbone when comparing with the memory-enhanced networks as in RVT \cite{Transformer4}. The comparison results for object detection are shown in Table \ref{tab:Gen1} and Table \ref{tab:1MPX}.

\begin{table}[t] %[htbp]\footnotesize
  \centering
    \tabcolsep=0.088cm
    \begin{adjustbox}{width=\columnwidth,center}
    \begin{tabular*}{\hsize}{@{}@{\extracolsep{\fill}}lccc@{}}
    & \multicolumn{3}{c}{Gen1} \\ 
    \toprule \toprule
    Methods
    & \multicolumn{1}{c}{mAP (\%)} 
    & \multicolumn{1}{c}{runtime (ms)} 
    & \multicolumn{1}{c}{Params (M)} \\  \midrule
    AEGNN \cite{GNN2}
    &16.3
    &- 
    &-  \\
    SNN-SSD \cite{SNN-SSD}
    &18.9
    &- 
    &8.2  \\
    SAM \cite{SAM}
    &35.5
    &- 
    &- \\
    FS \cite{framestack}
    &39.6$^\ast$
    &41.2 
    &- \\
    RED \cite{RED}
    & \ \ \ \ \ \  - / \textcolor{blue}{40.0}
    & \ \ \ \ \ \  - / \textcolor{blue}{16.7}
    & 24.1  \\
    ASTMNet \cite{ASTM}
    & 38.2 / \textcolor{blue}{46.7}
    & 28.6 / \textcolor{blue}{35.6}
    & 39.6  \\
    RVT-B \cite{Transformer4}
    % & \multicolumn{1}{|c}{Y} 
    & 32.0 / \textcolor{blue}{47.2}
    &  \ \ \ \ \ \  - / \textcolor{blue}{10.2}
    & 18.5  \\

    \midrule
    
    Swin-T v2 \cite{swinv2}
    & 34.3 / \textcolor{blue}{45.5}
    & 25.2$^\star$ / \textcolor{blue}{26.6$^\star$}
    & 17.6 / \textcolor{blue}{21.1}  \\
    Nested-T \cite{nest}
    & 35.1 / \textcolor{blue}{46.3}
    & 24.8$^\star$ / \textcolor{blue}{25.9$^\star$}
    & 18.7 / \textcolor{blue}{22.2}  \\
    Ours
    & \textbf{38.7} / \textcolor{blue}{\textbf{47.9}}
    & 15.9$^\star$ / \textcolor{blue}{16.8$^\star$}
    & 18.4 / \textcolor{blue}{21.9}  \\
    \bottomrule
    % \multicolumn{4}{l}{} \\
    \end{tabular*}%
    \end{adjustbox}
  \vspace{1mm}
  \caption{Detection performance on the Gen1 dataset. \textcolor{blue}{blue}$\colon$ The memory-enhanced results. $\ast\colon$ The result is in COCO AP50.
  $\star\colon$ The runtime includes event conversion time and inference time.}
  \vspace{-3mm}
  \label{tab:Gen1}%
\end{table}%

\begin{table}[t] %[htbp]\footnotesize
  \centering
    \tabcolsep=0.088cm
    \begin{adjustbox}{width=\columnwidth,center}
    \begin{tabular*}{\hsize}{@{}@{\extracolsep{\fill}}lccc@{}}
    & \multicolumn{3}{c}{1Mpx} \\ 
    \toprule \toprule
    Methods
    & \multicolumn{1}{c}{mAP (\%)} 
    & \multicolumn{1}{c}{runtime (ms)} 
    & \multicolumn{1}{c}{Params (M)} \\  \midrule
    SAM \cite{SAM} 
    &23.9
    &- 
    &-  \\
    UDA \cite{UDA}
    &48.0$^\ast$
    & - 
    & - \\
    RED \cite{RED}
    & 29.0 / \textcolor{blue}{43.0}
    & \ \ \ \ \ \  - / \textcolor{blue}{39.3}
    & 24.1  \\
    ASTMNet \cite{ASTM}
    & 40.3 / \textcolor{blue}{48.3}
    & 59.8 / \textcolor{blue}{72.3}
    & \textgreater 39.6  \\
    RVT-B \cite{Transformer4}
    % & \multicolumn{1}{|c}{Y} 
    & \ \ \ \ \ \  - / \textcolor{blue}{47.4}
    & \ \ \ \ \ \ - / \textcolor{blue}{11.9}
    & 18.5  \\

    \midrule
    
    Swin-T v2 \cite{swinv2}
    & 36.0 / \textcolor{blue}{46.4}
    & 33.6$^\star$ / \textcolor{blue}{34.5$^\star$}
    & 17.6 / \textcolor{blue}{21.1}  \\
    Nested-T \cite{nest}
    & 37.5 / \textcolor{blue}{46.0}
    & 32.1$^\star$ / \textcolor{blue}{33.5$^\star$}
    & 18.7 / \textcolor{blue}{22.2}  \\
    % Ours-
    % & 40.3 / \textcolor{blue}{47.7}
    % & 10.3 / \textcolor{blue}{11.3}
    % & \  \  8.3 / \textcolor{blue}{9.8}  \\
    Ours
    & \textbf{40.6} / \textcolor{blue}{\textbf{48.4}}
    & 17.1$^\star$ / \textcolor{blue}{18.2$^\star$}
    & 18.4 / \textcolor{blue}{21.9}  \\
    \bottomrule
    % \multicolumn{4}{l}{$\ast\colon$ the reported result is in COCO AP50, which is not comparable.} \\
    \end{tabular*}%
    \end{adjustbox}
  \vspace{1mm}
  %with the state-of-the-art methods 
  \caption{Detection performance on the 1Mpx dataset. \textcolor{blue}{blue}$\colon$ The memory-enhanced results.
  $\ast\colon$ The result is in COCO AP50.
  $\star\colon$ The runtime includes event conversion time and inference time.
  }
  \vspace{-3mm}
  \label{tab:1MPX}%
\end{table}%

\begin{figure*}[t]
\centering
\includegraphics[width=2\columnwidth]{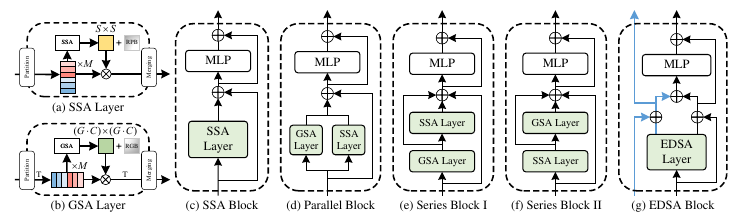}
\caption{
(a) The SSA layer. (b) The GSA layer. (c) The attention block with SSA operation only. (d) The attention block with both SSA and GSA operations, which are implemented in parallel. (e) The serial attention block performs the GSA operation first and then the SSA operation. (f) The serial attention block performs the SSA operation first and then the GSA operation. (g) The EDSA block contains dual residual connections.}
 \vspace{-3mm}
\label{fig:AS}
\end{figure*}

% \begin{table*}[t!] %[htbp]\footnotesize
% 	\centering
%         \begin{tabular*}{\hsize}{@{}@{\extracolsep{\fill}}lcccc|}
		
%         & \multicolumn{2}{c}{CIFAR10-DVS} 
%         & \multicolumn{2}{c}{Gen1} \\ 
%         \toprule \toprule
%         Methods
%         & \multicolumn{1}{c}{top-1 acc. (\%)} 
%         & \multicolumn{1}{c}{Param. (M)} 
%         & \multicolumn{1}{c}{mAP (\%)} 
%         & \multicolumn{1}{c}{runtime (ms)} \\ 
%         \midrule
        
%         Swin-T v2 
%         & \multicolumn{1}{c}{73.9} 
%         & \multicolumn{1}{c}{6.9} 
%         & \multicolumn{1}{c}{33.3} 
%         & \multicolumn{1}{c}{37.2} \\
  
% 		+ Group Token Embedding
% 		& \multicolumn{1}{c}{76.0\ (+2.1)} 
% 		& \multicolumn{1}{c}{7.2}
% 		& \multicolumn{1}{c}{34.9\ (+1.6)} 
% 		& \multicolumn{1}{c}{30.7} \\
  
% 		+ Event Dual Self-Attention (w/o GTA)
% 		& \multicolumn{1}{c}{79.4 \ (+5.5)} 
% 		& \multicolumn{1}{c}{7.6} 
% 		& \multicolumn{1}{c}{37.2 \ (+3.9)} 
% 		& \multicolumn{1}{c}{33.6} \\ 
  
%   	+ Group Token Aggregation (w/o EDSA)
% 		& \multicolumn{1}{c}{77.8 \ (+3.9)} 
% 		& \multicolumn{1}{c}{4.2}
% 		& \multicolumn{1}{c}{35.5\ (+2.2)} 
% 		& \multicolumn{1}{c}{24.6} \\
		
%   	+ GTA and EDSA
% 		& \multicolumn{1}{c}{\textbf{81.9} \ (+8.0)} 
% 		& \multicolumn{1}{c}{4.4} 
% 		& \multicolumn{1}{c}{\textbf{38.7} \ (+5.5)} 
% 		& \multicolumn{1}{c}{28.9} \\
%         \bottomrule
% 	\end{tabular*}%
%         \vspace{1mm}
% 	\caption{Performance of GET on the CIFAR10-DVS and Gen1 datasets with different modules.}
% 	\label{tab:ablation}%
% \end{table*}%

\begin{table*}[t!] %[htbp]\footnotesize
	\centering
        \begin{tabular*}{\hsize}{@{}@{\extracolsep{\fill}}lccccccc|}
	\multicolumn{4}{c}{}	
        & \multicolumn{2}{c}{CIFAR10-DVS} 
        & \multicolumn{2}{c}{1Mpx} \\ 
        \toprule \toprule
        Models
        &GTE
        & EDSA Block
        & GTA
        & \multicolumn{1}{c}{top-1 acc. (\%)} 
        & \multicolumn{1}{c}{Param. (M)} 
        & \multicolumn{1}{c}{mAP (\%)} 
        & \multicolumn{1}{c}{runtime (ms)} \\ 
        \midrule
        Model \uppercase\expandafter{\romannumeral1}
        &\scalebox{0.85}[1]{$\times$}
        & \scalebox{0.85}[1]{$\times$}
        & \scalebox{0.85}[1]{$\times$}
        & \multicolumn{1}{c}{77.2} 
        & \multicolumn{1}{c}{4.2} 
        & \multicolumn{1}{c}{37.5} 
        & \multicolumn{1}{c}{32.1} \\
  
        Model \uppercase\expandafter{\romannumeral2}
        &\checkmark
        & \scalebox{0.85}[1]{$\times$}
        & \scalebox{0.85}[1]{$\times$}
		& \multicolumn{1}{c}{79.9\ (+2.7)} 
		& \multicolumn{1}{c}{4.4}
		& \multicolumn{1}{c}{39.1\ (+1.6)} 
		& \multicolumn{1}{c}{16.2} \\
  
        Model \uppercase\expandafter{\romannumeral3}
        &\checkmark
        & \checkmark
        & \scalebox{0.85}[1]{$\times$}
		& \multicolumn{1}{c}{81.5\ (+4.3)} 
		& \multicolumn{1}{c}{4.7} 
		& \multicolumn{1}{c}{40.0\ (+2.5)} 
		& \multicolumn{1}{c}{18.2} \\ 
  
        Model \uppercase\expandafter{\romannumeral4}
        &\checkmark
        & \scalebox{0.85}[1]{$\times$}
        & \checkmark
		& \multicolumn{1}{c}{79.4\ (+2.2)} 
		& \multicolumn{1}{c}{4.2}
		& \multicolumn{1}{c}{39.3\ (+1.8)} 
		& \multicolumn{1}{c}{14.9} \\
		
        GET
        &\checkmark
        & \checkmark
        & \checkmark
		& \multicolumn{1}{c}{\textbf{84.8}\ (+7.6)} 
		& \multicolumn{1}{c}{4.5} 
		& \multicolumn{1}{c}{\textbf{40.6}\ (+3.1)} 
		& \multicolumn{1}{c}{17.1} \\
        \bottomrule
	\end{tabular*}%
        \vspace{1mm}
	\caption{Performance of GET on the CIFAR10-DVS and 1Mpx datasets with different modules. \scalebox{0.85}[1]{$\times$}$\colon$ modules of Nested-T is used.}
 \vspace{-2mm}
	\label{tab:ablation}%
\end{table*}%

\begin{table*}[t!] %[htbp]\footnotesize
	\centering

        \begin{tabular*}{\hsize}{@{}@{\extracolsep{\fill}}lccccc@{}}
	
        \toprule \toprule
        
        Representations
        & \multicolumn{1}{c}{Group Token}
        & \multicolumn{1}{c}{Event Histogram \cite{HIST}}
        & \multicolumn{1}{c}{Voxel Grid \cite{VOXEL}}
        & \multicolumn{1}{c}{Time Surface \cite{HOTS}}
        & \multicolumn{1}{c}{TORE \cite{tore}}\\
        \midrule
        
        top-1 acc. (\%)
        & \multicolumn{1}{c}{\textbf{84.8}}
        & \multicolumn{1}{c}{75.8}
        & \multicolumn{1}{c}{78.4}
        & \multicolumn{1}{c}{80.2}
        & \multicolumn{1}{c}{80.4}
        \\  
        
        Time (s / $10^{8}$evs)
	& \multicolumn{1}{c}{\textbf{0.052}}
        & \multicolumn{1}{c}{0.374}
        & \multicolumn{1}{c}{0.390}
        & \multicolumn{1}{c}{4.235}
        & \multicolumn{1}{c}{16.027}
        \\  
        \bottomrule
        % & \multicolumn{5}{l}{Classification performance on the DVS128Gesture dataset.}
	\end{tabular*}%
  
        \vspace{1mm}
        
	\caption{Classification performance on the Cifar10-DVS dataset, and the time cost of converting $10^8$ events into various representations.}
 \vspace{-3mm}
	\label{tab:representations}%
\end{table*}%

\begin{table}[t!] %[htbp]\footnotesize
	\centering
        \begin{tabular*}{\hsize}{@{}@{\extracolsep{\fill}}lcccc@{}}
		
        & \multicolumn{2}{c}{CIFAR10-DVS} \\ 
        \toprule \toprule
        Methods
        & \multicolumn{1}{c}{top-1 acc. (\%)}
        & \multicolumn{1}{c}{Param. (M)} \\
        \midrule
        
        SSA Block
	& \multicolumn{1}{c}{79.4}
        & \multicolumn{1}{c}{4.2}  \\  

        Parallel Block
	& \multicolumn{1}{c}{78.5}
        & \multicolumn{1}{c}{4.5}  \\ 
  
        Series Block I
	& \multicolumn{1}{c}{75.0}
        & \multicolumn{1}{c}{4.5}  \\   
  
        Series Block II
	& \multicolumn{1}{c}{81.9} 
        & \multicolumn{1}{c}{4.5}  \\ 
  
	EDSA Block
	& \multicolumn{1}{c}{\textbf{84.8}} 
        & \multicolumn{1}{c}{4.5}  \\
        \bottomrule
	\end{tabular*}%
        \vspace{1mm}
	\caption{Classification performance of GET on the CIFAR10-DVS dataset with different self-attention blocks.}
 \vspace{-3mm}
	\label{tab:attention ablation}%
\end{table}%

Table \ref{tab:Gen1} shows that GET performs best on the Gen1 dataset. The state-of-the-art network without using any memory mechanism is ASTMNet, with an mAP of $38.2\%$. GET performs better at $38.7\%$, while the parameter number is less than its $47\%$. It also maintains an optimal mAP of $47.9\%$ when using the memory mechanism, which is $0.7\%$ larger than the RVT-B of $47.2\%$. Compared with Nested-T (with YOLOX framework, using patched voxel grid as event representation), the gains with and without memory mechanism are $3.6\%$ and $1.6\%$ respectively. GET also achieves the optimal result on the 1Mpx dataset, as shown in Table \ref{tab:1MPX}. When the memory mechanism is not used, GET outperforms the state-of-the-art network ASTMNet ($40.6 \%$ vs. $40.3 \%$). With the help of ConvLSTM layers, GET achieves the highest mAP of $48.4 \%$, compared with ASTMNet of $48.3 \%$. Compared with Nested-T, the gains with and without memory mechanism are $3.1\%$ and $2.4\%$. 

On both Gen1 and 1Mpx datasets, GET is the fastest end-to-end method. Its combined runtime is even smaller than the data preprocessing time of other methods.

\subsection{Ablation Studies}
To analyze the proposed methods for GET, we conduct a series of ablative experiments on the CIFAR10-DVS, DVS128Gesture, and 1Mpx datasets. We construct four variants Model I-IV, which have different combinations of proposed modules. Table \ref{tab:ablation} reports the performance of these variants and GET.

\noindent\textbf{Group Token Embedding.}
Group Token enriches patch temporal-polarity information, yielding a 2.7\% increase in top-1 accuracy and a 1.6\% rise in mAP. Furthermore, the implementation of this approach has resulted in a 51\% reduction in runtime. Hence, Group Token is an efficient and effective method to improve performance by enhancing temporal-polarity information.

We also conduct experiments to convert $10^8$ events to different representations. As shown in Group Token Embedding \ref{tab:representations}, our scheme only requires only 0.052s, which is much less than using other methods' officially released codes. For performance comparison, we use VIT’s \cite{VIT} patch embedding module for tokenizing other representations. It can be observed that our proposed Group Token outperforms others by a large margin ($84.8 \%$ vs. $80.4 \%$). 

\noindent\textbf{Event Dual Self-Attention.} 
EDSA is an effective approach to extract features from both the spatial and temporal-polarity domains while maintaining cross-pixel and cross-group communications. In comparison with the commonly used SSA approach, EDSA improves the top-1 accuracy and mAP by 1.6\% and 0.9\%, respectively. However, there is a slight increase in the number of parameters and running time. To demonstrate the superiority of EDSA's block structure, we compared it with other possible block structures, as illustrated in Figure \ref{fig:AS}. Table \ref{tab:attention ablation} presents the performance of GET using these attention blocks, indicating that EDSA block with dual residual connections achieves the best result ($84.8 \%$ vs. $81.9 \%$). Therefore, EDSA is a promising approach for feature extraction in computer vision tasks.

\noindent\textbf{Group Token Aggregation.} 
% GTA increases the top-1 accuracy and mAP by $1.8\%$ and $0.6\%$. Leveraging the overlapping group convolution design, GTA achieves a reliable integration of spatial and temporal-polarity information at a shallower stage (three). As a result, the number of parameters is greatly reduced from 7.2 M to 4.2 M, and the running time is reduced by 6.1 ms.
% The combination of Group Token Embedding, Event Dual Self-Attention, and Group Token Aggregation leads to the best top-1 accuracy and mAP of $81.9\%$ and $38.7\%$. 
GTA leverages overlapping group convolution to decouple spatial and temporal-polarity information at an earlier stage. The resulting model is more parameter-efficient, with parameters reduced by 0.2 M, and faster, with a 1.3 ms reduction in running time. However, without EDSA to compensate for the reduced inter-channel interaction caused by group convolution, GTA will lead to a slight decrease in performance. 

\noindent\textbf{GET.} 
When combined with the GTE module, EDSA block, and GTA module, GET achieves the highest top-1 accuracy and mAP of 84.8\% and 40.6\%.

\begin{table}[t!] %[htbp]\footnotesize
	\centering
        \tabcolsep=0.088cm
        \begin{tabular*}{\hsize}{@{}@{\extracolsep{\fill}}lccccc@{}}
		
        & \multicolumn{3}{c}{CIFAR10-DVS} \\ 
        \toprule \toprule
        Modules / Models
        & \multicolumn{1}{c}{Swin-T v2 \cite{swinv2}}
        & \multicolumn{1}{c}{Nested-T \cite{nest}}
        & \multicolumn{1}{c}{GET}\\
        \midrule
        
        Token Embedding
	& \multicolumn{1}{c}{78.4}
        & \multicolumn{1}{c}{78.4}
        & \multicolumn{1}{c}{\textbf{84.8}}  \\  

        Self-Attention Block
	& \multicolumn{1}{c}{79.4}
        & \multicolumn{1}{c}{79.8}
        & \multicolumn{1}{c}{\textbf{84.8}}  \\ 
  
        Token Aggregation
	& \multicolumn{1}{c}{80.4}
        & \multicolumn{1}{c}{81.0}
        & \multicolumn{1}{c}{\textbf{84.8}}  \\   

        \bottomrule
	\end{tabular*}%
        \vspace{1mm}
	\caption{Module substitution comparison on the CIFAR10-DVS dataset. The number indicates the performance of a GET variant with a specific module replaced by the corresponding module of Swin-T v2 and Nested-T.}
  \vspace{-3mm}
 \label{tab:module ablation}%
\end{table}

\begin{table}[t!]
    \centering
    \tabcolsep=0.058cm
    \begin{tabular*}{\hsize}{@{}@{\extracolsep{\fill}}lccccc@{}}
        Datasets / \#Group & 4 & 6 & 12 & 24 & 48\\
        \toprule \toprule
        \multicolumn{1}{l}{Cifar10-DVS} & 79.9 & 81.1 & \textbf{84.8} & 82.3 & 77.6\\
        \multicolumn{1}{l}{DVS128Gesture} & 94.8 & 95.8 & 97.2 & \textbf{97.9} & 96.9\\
        % \midrule
        % \multicolumn{1}{l}{Param. (M)} & 4.4 & 4.4 & 4.5 & 4.6 & 4.7\\
        \bottomrule
        \end{tabular*}
        \vspace{1mm}
    \caption{Classification performance of GET on the Cifar10-DVS and DVS128Gesture datasets with different group numbers.}
    \vspace{-3mm}
    \label{tab:group number ablation}
\end{table}

\noindent\textbf{Module Superiority.} To demonstrate the superiority of our proposed modules, we compare them with corresponding modules from other networks. The results in Table \ref{tab:module ablation}, show that our modules outperform the rest on the CIFAR10-DVS dataset. Specifically, the token embedding modules of Swin-T v2 and Nested-T both embed voxel grids into patches and achieve an accuracy of $78.4\%$. The self-attention modules of the two models (SSA Block in Figure \ref{fig:AS}, w/ and w/o shift-window operation) result in $79.4\%$ and $79.8\%$. And the token aggregation modules of Swin-T v2 and Nested-T achieve $80.4\%$ and $81.0\%$ accuracy by using patch merging and conv-pool methods. Our proposed modules outperform all with an accuracy of $84.8\%$.

\noindent\textbf{Group Number.} 
Table \ref{tab:group number ablation} provides the performance on the Cifar10-DVS and DVS128Gesture datasets when choosing different group numbers $G$. The Cifar10-DVS dataset contains short-time-range event streams. As a result, when $G$ is chosen as 12 ($2K=24$), GET accurately classifies most objects and results in an $84.8\%$ top-1 accuracy. As $G$ increases, performance deteriorates because events in each group become sparse. The DVS128Gesture dataset contains long-time-range event streams. If $G$ is small, the action overlaps within a group, degrading performance. When $G$ is chosen as 24 ($2K=48$), we get the highest accuracy of $97.9\%$.

% \begin{table}[t!] %[htbp]\footnotesize
% 	\centering
%         \tabcolsep=0.058cm
%         \begin{tabular*}{\hsize}{@{}@{\extracolsep{\fill}}cccccc@{}}
		
%         & \multicolumn{1}{c}{Cifar10-DVS}
%         & \multicolumn{1}{c}{DVS128Gesture}\\ 
%         \toprule \toprule
%         \#Group
%         & \multicolumn{2}{c}{top-1 acc. (\%)}
%         & \multicolumn{1}{c}{Param. (M)} \\
%         \midrule
        
%         4
%         & \multicolumn{1}{c}{80.8}
% 	& \multicolumn{1}{c}{94.6}
%         & \multicolumn{1}{c}{4.4}  \\  

%         6
%         & \multicolumn{1}{c}{\textbf{81.9}} 
% 	& \multicolumn{1}{c}{95.2}
%         & \multicolumn{1}{c}{4.4}  \\ 
  
%         12
%         & \multicolumn{1}{c}{81.3}
% 	& \multicolumn{1}{c}{97.0}
%         & \multicolumn{1}{c}{4.5}  \\   
  
%         24
%         & \multicolumn{1}{c}{79.9}
% 	& \multicolumn{1}{c}{\textbf{97.7}} 
%         & \multicolumn{1}{c}{4.6}  \\ 
  
% 	48
%         & \multicolumn{1}{c}{78.9}
% 	& \multicolumn{1}{c}{96.4} 
%         & \multicolumn{1}{c}{4.7}  \\
%         \bottomrule
% 	\end{tabular*}%
%         \vspace{1mm}
% 	\caption{Classification performance of GET on the Cifar10-DVS and DVS128Gesture datasets with different group numbers.}
% 	\label{tab:group number ablation}%
% \end{table}%
\begin{figure}[t]
\centering
\includegraphics[width=\columnwidth]{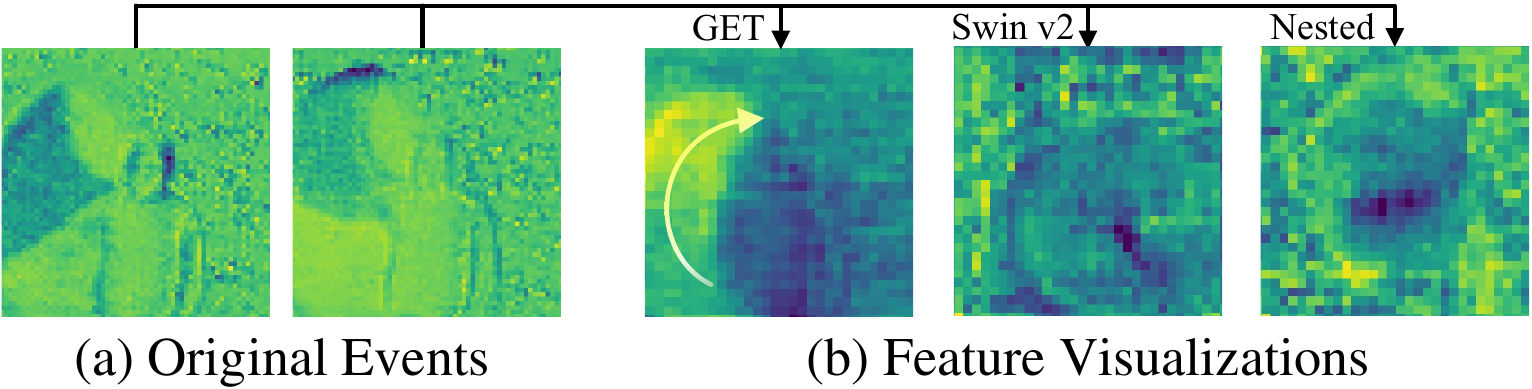}
\caption{
(a) Visualizations of original events (right-hand counterclockwise waving in DVS128Gesture dataset). (b) Feature visualizations of different Transformers utilizing EDSA and SSA.}
 \vspace{-3mm}
\label{fig:FV}
\end{figure}

\begin{figure}[t]
\centering
\includegraphics[width=\columnwidth]{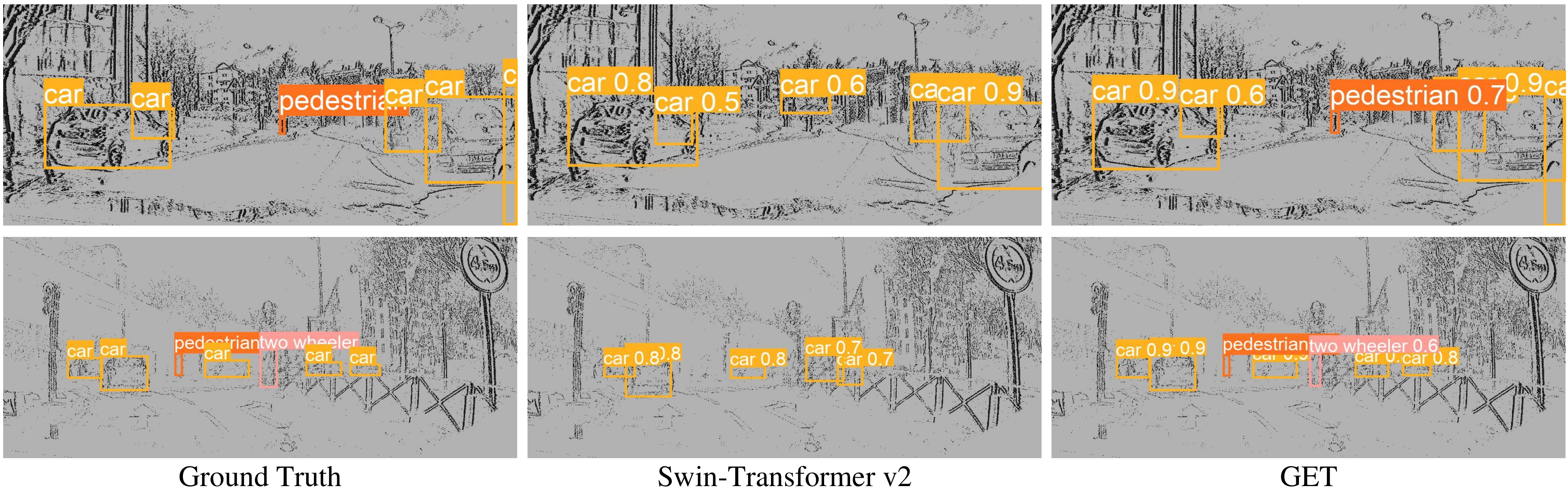}
\caption{
 Visualizations of the detection results on the 1Mpx dataset. Compared with the Swin-T v2, GET detects most objects, while the bounding boxes are more accurate.}
\label{fig:Detectionresults}
\vspace{-3mm}
\end{figure}

\subsection{Visualizations}
we present visualizations of the feature maps extracted by GET and other Transformers in Figure \ref{fig:FV}.
It can be observed that only GET, through the utilization of EDSA, successfully captures counterclockwise motion. This becomes evident as we compare the results. The other two classifiers that use SSA as the token mixer inaccurately categorized the sample as right-hand clockwise waving.

% Among the three classifiers, only GET can correctly recognize the sample as right-hand counterclockwise waving.
% It can be observed that the local correlations of different times and polarities are preserved in small blocks in the spatial dimension. 
% With EDSA, these temporal-polarity correlations can be used to refine the spatial correlations represented by the feature in Figure \ref{fig:FV} (b) and result in the feature shown in Figure \ref{fig:FV} (d).
We also present object detection results on the 1Mpx dataset in Figure \ref{fig:Detectionresults}. It can be seen that GET detects the most objects compared with Swin-T v2, especially when objects are intertwined with the background. The bounding boxes detected by GET are also more accurate.

\section{Conclusion}
In this paper, we propose a novel event-based vision backbone, called Group Event Transformer (GET), that effectively utilizes the temporal and polarity information of events while preserving the fair spatial modeling capability of traditional Transformer-based backbones. The proposed backbone incorporates the Group Token representation, the EDSA block, and the GTA module, which enable effective feature communication and integration in both the spatial and temporal-polarity domains. The experimental results on four event-based classification datasets
(Cifar10-DVS, N-MNIST, N-CARS, and DVS128Gesture) and two event-based object detection datasets (1Mpx and Gen1) all show that our method achieves superior performance over state-of-the-art methods.

\section{Acknowledgement}
This work is in part supported by the National Key R\&D Program of China under Grant 2020AAA0108600 and the National Natural Science Foundation of China under Grant 62032006.
% \pagebreak
% \newpage

{\small
\bibliographystyle{ieee_fullname}
\bibliography{egbib}
}

\end{document}